\documentclass[runningheads]{llncs}
\usepackage{graphicx}
\usepackage{amsmath} 
\usepackage{mathtools}

\usepackage{graphicx}
\usepackage{caption}
\usepackage{subfig}
\usepackage{subfig}
\usepackage{tabularx}

\usepackage{algorithm}
\usepackage[noend]{algpseudocode}

\usepackage{multirow}
\begin{document}
\title{Multiple Object Tracking in Urban Traffic Scenes with a Multiclass Object Detector
}
\titlerunning{MOT in Urban Traffic Scenes with a Multiclass Object Detector}
%
\authorrunning{Ooi et al.}
\author{Hui-Lee Ooi 
\and
Guillaume-Alexandre Bilodeau
\and
Nicolas Saunier
\and David-Alexandre Beaupr\'{e}
}
%
\institute{Polytechnique Montr\'{e}al, Canada \\ \email{<hui-lee.ooi, gabilodeau, nicolas.saunier, \\ david-alexandre.beaupre>@polymtl.ca}
}

\maketitle              
\begin{abstract}
Multiple object tracking (MOT) in urban traffic aims to produce the trajectories of the different road users that move across the field of view with different directions and speeds and that can have varying appearances and sizes. Occlusions and interactions among the different objects are expected and common due to the nature of urban road traffic. In this work, a tracking framework employing classification label information from a deep learning detection approach is used for associating the different objects, in addition to object position and appearances. We want to investigate the performance of a modern multiclass object detector for the MOT task in traffic scenes. Results show that the object labels improve tracking performance, but that the output of object detectors are not always reliable.

\end{abstract}

\section{Introduction}
\begin{figure}
\centering
\includegraphics[width=0.5\linewidth]{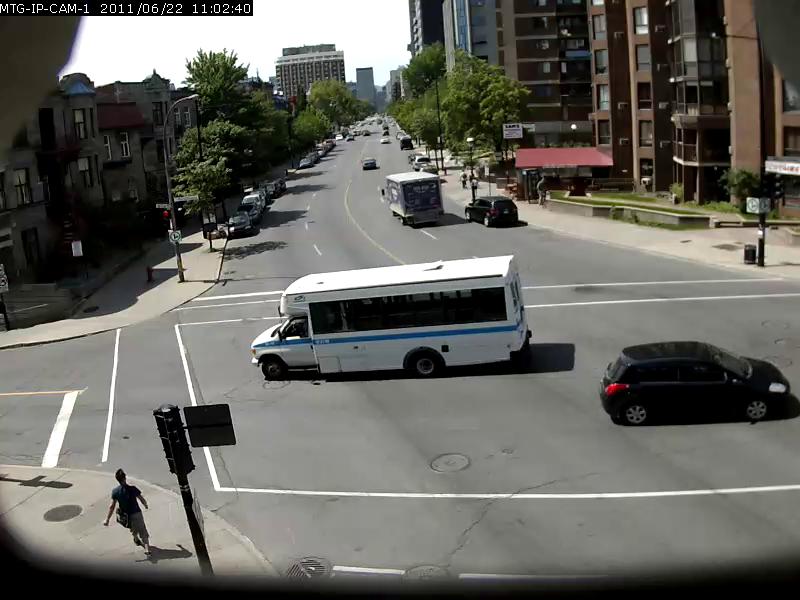}
\caption{A frame from the urban traffic dataset that shows several road users in an intersection.} \label{fig:urban}
\end{figure}

The objective of multiple object tracking (MOT) is extracting the trajectories of the different objects of interest in the scene (camera field of view). It is a common computer vision problem that is still open in complex applications. This paper deals with one of these complex applications, urban traffic, that involves different kinds of road users such as drivers of motorized and non-motorized vehicles, and pedestrians (see Figure \ref{fig:urban}). The various road users exhibit different properties of moving speeds and directions in the urban environment. Their size vary because of perspective. Besides, road users are frequently interacting and occluding each other, which makes it even more challenging. 

In this work, we want to investigate the performance of a modern multiclass object detector \cite{ref_rfcn} for the MOT task in traffic scenes. We are interested in testing MOT in urban traffic settings with road users of varying sizes using an object detector while most previous works in such applications employ background subtraction or optical flow to extract the objects of interest regardless of their size. Our contributions in this work is an assessment of a typical model object detector for tracking in urban traffic scenes, and the introduction of label information for describing the objects in the scenes. Due to the variability of objects found in urban scenes, the label information should be a useful indicator to distinguish and associate the objects of interests across frames, thereby producing a more accurate trajectory. In this paper, the improvements obtained thanks to classification labels are evaluated with respect to a baseline tracker that uses a Kalman filter, bounding box positions and color information.

The results show that using classification labels from a detector improves significantly tracking performances on an urban traffic dataset. Therefore, multiple object trackers should capitalize on this information when it is available. However, they also show that the outputs of a multiclass object detector are not always reliable and not always easy to interpret. 

\section{Related Works}
MOT in urban traffic scenes was previously studied in \cite{ref_urbantracker}, where the use of background subtraction is proposed for detecting the objects of interest followed by updating the object model with a state machine that uses feature points and spatial information. In fact, most previous work in MOT uses background subtraction or optical flow to detect the objects. The reason is that historically, methods based on pre-trained bounding box detectors are difficult to apply to road user tracking scenarios because it is difficult to design a detector that can detect and classify every possible type of road user from various viewpoints. However, recent progress in deep learning \cite{ref_rfcn,ref_yolo} make this avenue now possible and worth investigating. 

When using background subtraction, the detection results give blobs that can correspond to parts of objects, one object, or many objects grouped together. The task is then to distinguish between merging, fragmentation, and splitting of objects. This is the main drawback of this method, since under congested traffic conditions, road users may partially occlude each other and therefore be merged into a single blob. Examples of trackers based on background subtraction include the work of Fuentes and Velastin \cite{ref_Fuentes2006}, Torabi et al.\ \cite{ref_Torabi2009}, Jun et al.\ \cite{ref_jun_tracking_2008}, Kim et al. \cite{ref_kim_real_2008}, Mendes et al.\ \cite{ref_mendes15vehicle}, and Jodoin et al. \cite{ref_JodoinITS2016}. For data association, they typically use the overlap of foreground blobs between two frames or a graph-based model for data association using appearance information, such as textures, color or keypoints. These approaches track objects in a merge-split manner as objects are tracked as groups during occlusion. The Hungarian algorithm is a classical graph-based choice for solving object assignment problems. To compensate for the missing detections, the Kalman filter is a popular option for estimating the location of the object of interest. A basic implementation of multiple object tracking is proposed in \cite{ref_sort} using this approach. 

With optical flow, objects are detected by studying the motion of tracked points in a video. Feature points that are moving together belongs to the same object. Several methods accomplish this process using the Kanade-Lucas-Tomasi (KLT) tracker \cite{ref_KLT}. The following researchers have proposed such trackers, often called feature-based: Beymer et al.\ \cite{ref_Beymer1997}, Coifman et al.\ \cite{ref_coifman_real-time_1998}, Saunier et al.\ \cite{ref_saunier_feature_based_2006} and Aslani and Mahdavi-Nasab \cite{ref_Aslani2013}. For example, the algorithm proposed by Saunier et al. \cite{ref_saunier_feature_based_2006}, named Traffic Intelligence, tracks road users at urban intersections by continuously detecting new features. The main issue is to select the right parameters to segment objects moving at similar speeds, while at the same time not oversegmenting smaller non-rigid objects such as pedestrians. Because objects are identified only by their motion, nearby road users moving at similar speed are often merged together. The exact bounding box occupied by the road user is unknown because it depends on the position of sparse feature points. Furthermore, when an object stops, its features flow becomes zero and feature trajectories are interrupted, which leads to fragmented object trajectories. Using a deep learning-based detector on road users is expected to provide objects that are less fragmented and that can be tracked whether they are moving or not.  

\section{Method}
The proposed method consists of two main components: object detection and data association. It is illustrated in Algorithm \ref{pseudocode}. Object detection involves the extraction of objects of interest from the frames for further processing. Data association determines the tracking architecture to ensure the formation of the trajectories of each object in the scene. 
In order to match the objects correctly, an assignment cost based on a measure of similarity is computed for all the potential matches. 

\subsection{Object Detection}
The road users from each frame are detected by using a deep-learning object detection model from the Region-based Fully Convolutional Network (RFCN) \cite{ref_rfcn} framework due to its efficiency and accuracy. This detector was selected because it was the best performing approach on the MIO-TCD localization challenge \cite{ref_miotcd}. The pre-trained model is further refined by using the MIO-TCD dataset \cite{ref_miotcd} to provide the labels of the different road users found in traffic scenes, belonging to one of the eleven categories or labels: articulated truck, bicycle, bus, car, motorcycle, motorized vehicle, non-motorized vehicle, pedestrian, pickup truck, single unit truck and work van.  

A non-maximal suppression (NMS) method \cite{ref_ensemble,ref_objdetection} is applied to reduce the redundant detections of the same road users in each frame. 

\subsection{Data Association} 
The object assignment or data association is essentially performed on a set of detected objects from the current frame and a list of actively tracked objects that are accumulated from previous frames. 

For the matched pairings, the latest position of the corresponding object in the track list is updated from the detected object. In the case of new detection, a new object will be initialized and added to the track list. In the case of objects in the track list without a matched candidate from the detection list, i.e.\ a missing detection, a Kalman filter \cite{ref_kalman} is applied to predict its subsequent location in the scene and the track information is updated using the prediction.

For the matching of objects across frames, if the total cost of assigning object pairs is higher than a set threshold $T_{match}$, the paired object would be reassigned to unmatched detection and unmatched track respectively due to the high probability of them not being a good match. 

\begin{algorithm}
\caption{MOT algorithm}\label{pseudocode}
\begin{algorithmic}[1]
\Procedure{MOT}{}
\For{\texttt{$i^{th}$ frame}}
	\State
   	Extract detections with multiclass object detector       
    \If {$i ==1$} 
    	\State Assign all detections as tracks
    \Else
        \For{each detection}
        	\State Compute cost of detection with respect to each track
      	\EndFor
        \State Run Hungarian algorithm for assigning pairing of detection and track 
      	\For{each matched detection}
        	\If {$Cost> T_{match}$} 
        		\State Reassign as unmatched detection and unmatched track
        	\Else
        		\State Update the track information from the detection
	   		\EndIf
      	\EndFor
      	\For{each unmatched detection}
        	\State Initialize as new track
      	\EndFor
      	\For{each unmatched track}
        	\If {$N > N_{timeout}$} 
        		\State Remove track
        	\Else
        		\State Update track information using prediction from Kalman filter
			\EndIf
      	\EndFor     
	\EndIf
\EndFor
\EndProcedure
\end{algorithmic}
\end{algorithm}

Actively tracked objects that are not assigned a corresponding object from the new detections after $N_{timeout}$ frames are removed from the list, under the assumption that the object has left the scene or the object was an anomaly from the detection module. 

\paragraph{Object Assignment Cost} 
Once objects are detected, the subsequent step is to link the correct objects by using sufficient information about the objects to compute the cost of matching the objects. The Hungarian algorithm \cite{ref_hungarian} is applied to match the list of active objects with the list of new detections in the current frame so that the matchings are exclusive and unique. The bipartite matching indicates that each active object can only be paired with one other candidate object (the detection) from the current frame. 
The algorithm can make use of different costs of assignment, with higher costs given to objects that are likely to be different road users. 

\paragraph{Label Cost} 
In order to describe the properties of the detected objects, the labels and corresponding confidence score from the detections are taken into account. Setting the range of scores between 0 and 1, object pairs across frames that are more similar will be given a lower cost. Using the classification labels, object pairs with different labels are less likely to be the correct matchings, therefore they will be given cost of 1. Meanwhile, when the pairing labels are the same, the average of the confidence score of each detection are being taken as the label cost. The label cost is defined as

\begin{equation}
  C_{label}=
  \begin{cases}
    1 - 0.5\times($Conf$_{i} + $Conf$_{j}) & \text{if $L_{i}=L_{j}$ } \\
    1 & \text{if $L_{i} \neq L_{j}$} 
  \end{cases}
\label{eqn:cost_label}
\end{equation}
where $L_{n}$ denotes the label of detection $n$ and $Conf_{n}$ denotes the confidence of the corresponding label of the $n^{th}$ detection.

\paragraph{Jaccard Distance-based Position Cost}
The bounding box coordinates of the detected objects are a useful indicator for matching the objects across frames as well. To judge the similarity of two bounding boxes in terms of proximity and size, the Jaccard distance is computed from the coordinates of the paired object, where the ratio of intersection over union of the bounding boxes is computed. This is calculated using

\begin{equation}
  C_{position} = 1- \frac{|Box_i \cap Box_j|}{|Box_i \cup Box_j|}
  \label{eqn:cost_pos}
\end{equation}
where $Box_{n}$ denotes the set of pixels of the bounding box of the detected object $n$.

\paragraph{Color Cost} 
The visual appearance of the objects is characterized by their color histograms that are used to compute the color cost. In this work, the Bhattacharyya distance is applied to compute the distance of the color histogram of detections across frames with

\begin{equation}
  C_{color} =\sqrt[]{1-\frac{1}{\sqrt[]{\bar{H_{i}}\bar{H_{j}}N^2}}\sum{\sqrt[]{H_i H_j}}} 
  \label{eqn:cost_color}
\end{equation}
where $H_{i}$ denotes the color histogram of detection $i$, $H_{j}$ denotes the color histogram of detection $j$ and $N$ is the total number of histogram bins.

\section{Results and Discussion}
\begin{figure}[ht!]
\centering
\begin{tabular}{cc}
\subfloat{\includegraphics[width=0.49\textwidth]{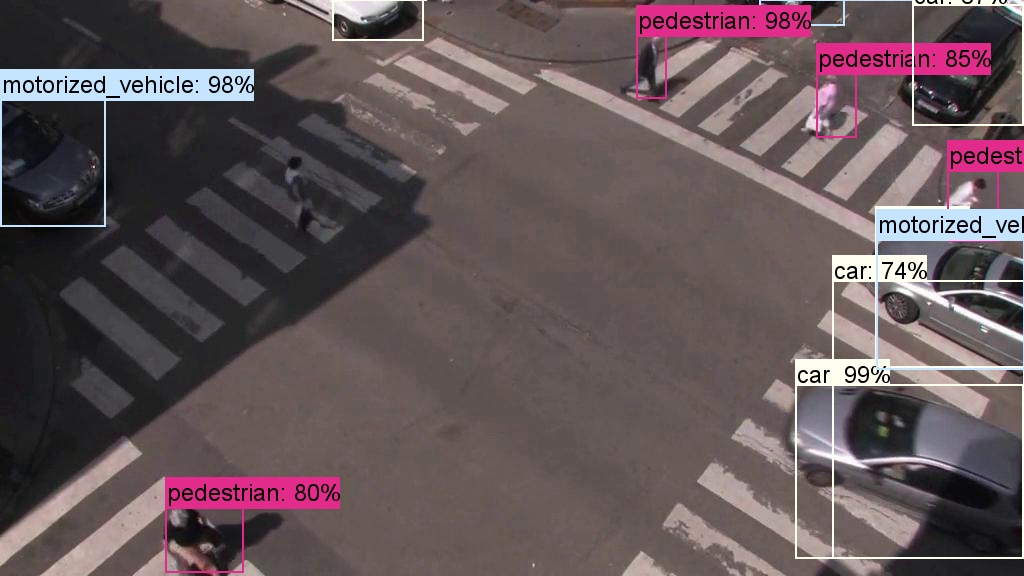}}
& \subfloat{\includegraphics[width=0.49\textwidth, trim={0 3.5cm  0 1.5cm},clip]{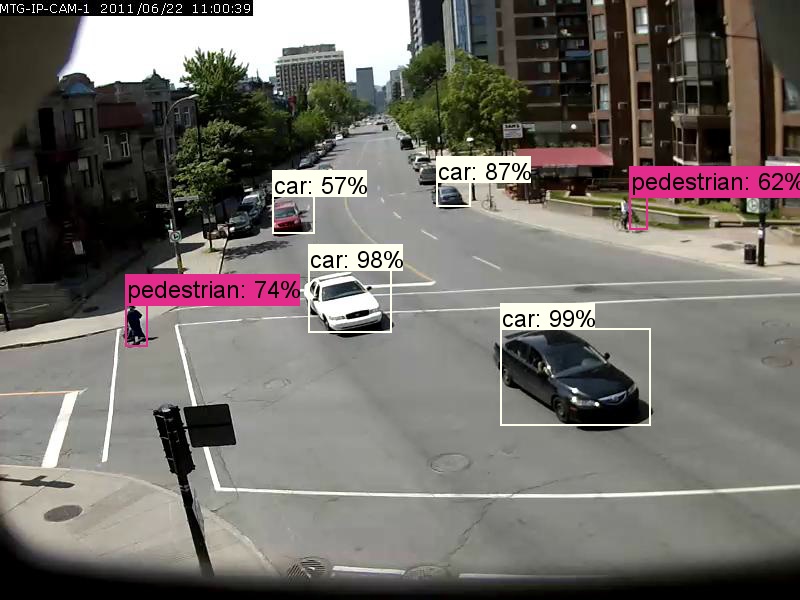}}\\
\subfloat{\includegraphics[width=0.49\textwidth]{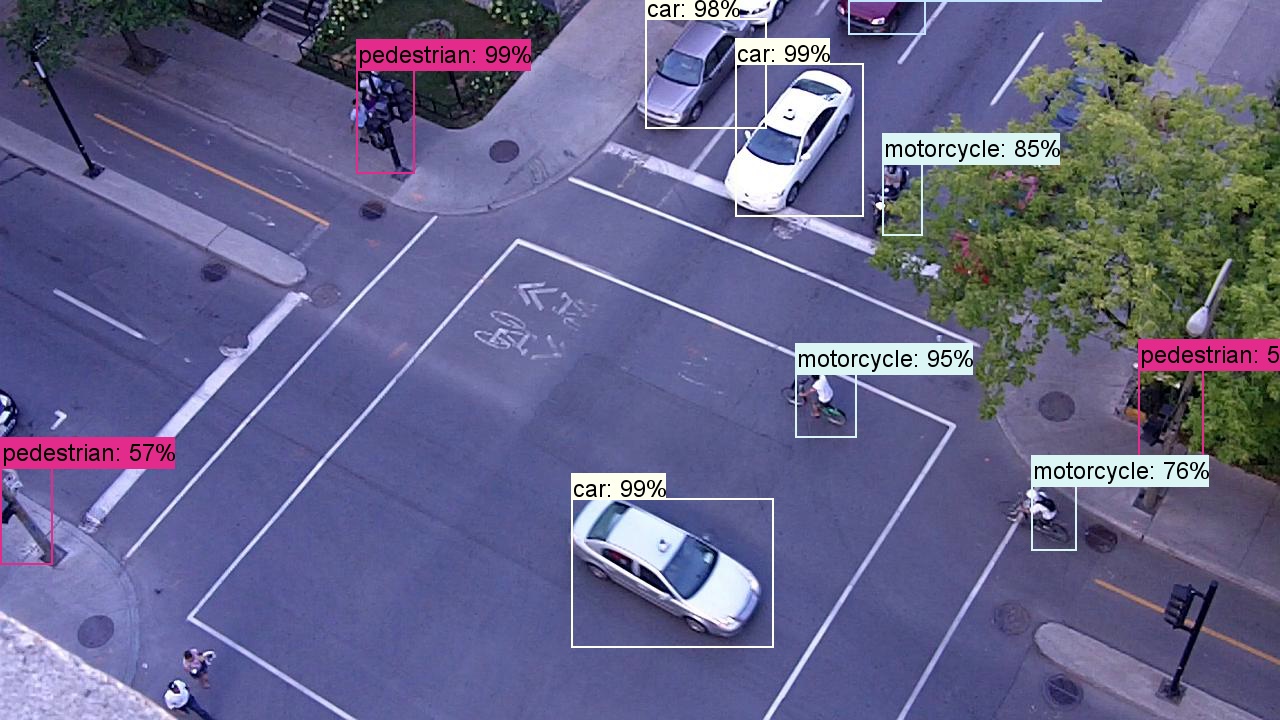}}
& \subfloat{\includegraphics[width=0.49\textwidth]{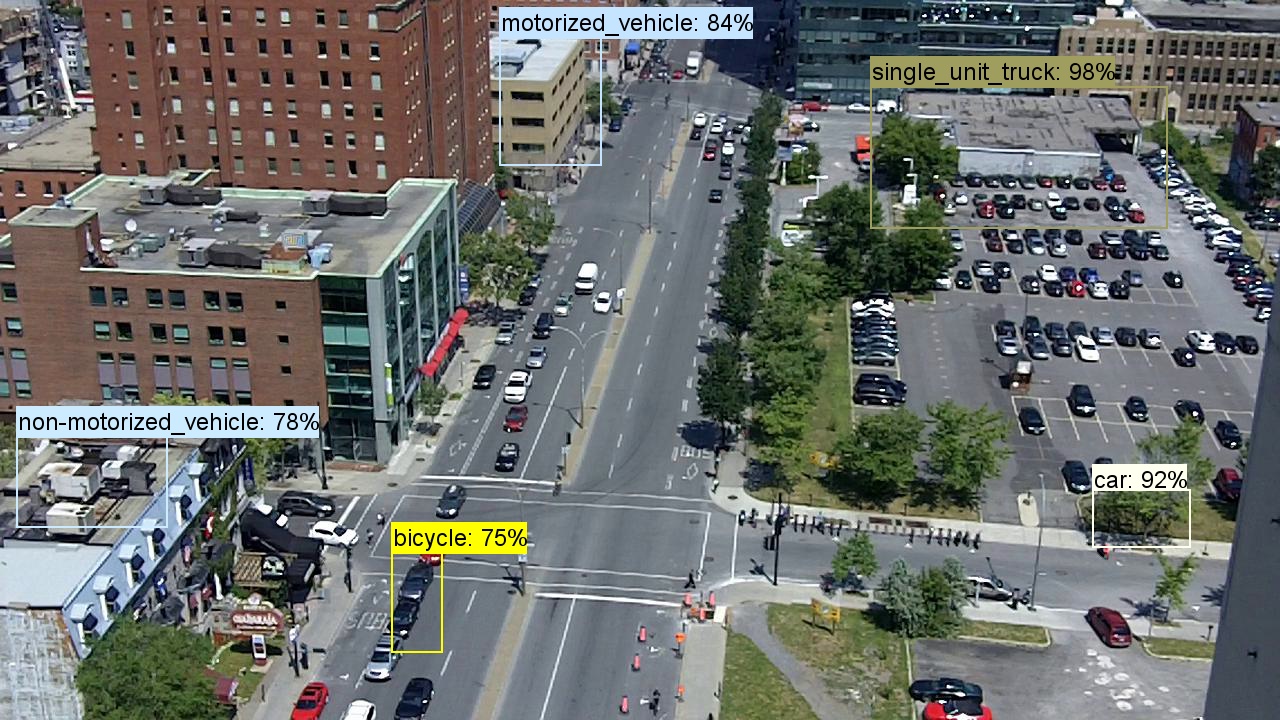}}\\
\end{tabular}
\caption{Samples frames with detections from the Urban Tracker dataset}
\label{samples}
\end{figure}

To test the RFCN multiclass object detector in MOT and to assess the usefulness of the classification labels, we used the Urban Tracker dataset \cite{ref_urbantracker} since it contains a variety of road users in an urban environment. Fig~\ref{samples} shows some sample frames from the Urban Tracker dataset with RFCN detections. The MOT performance is evaluated by using the CLEAR MOT metrics \cite{ref_clearmot}: 
\begin{itemize}
\item multiple object tracking accuracy (MOTA) that evaluates the quality of the tracking, if all road users are correctly detected in the frames they are visible and if there are no false alarms;
\item and multiple object tracking precision (MOTP) that evaluates the quality of the localization of the matches.
\end{itemize}

To test the contribution of using labels in MOT, the proposed baseline method is applied with and without object classification labels in the cost computation for data association. The following parameters are used in the experiments: $T_{match}$ is set at 1.5 and the value of $N_{timeout}$ is set at 5. 

\begin{figure}[ht!]
\centering
\includegraphics[width=0.6\linewidth]{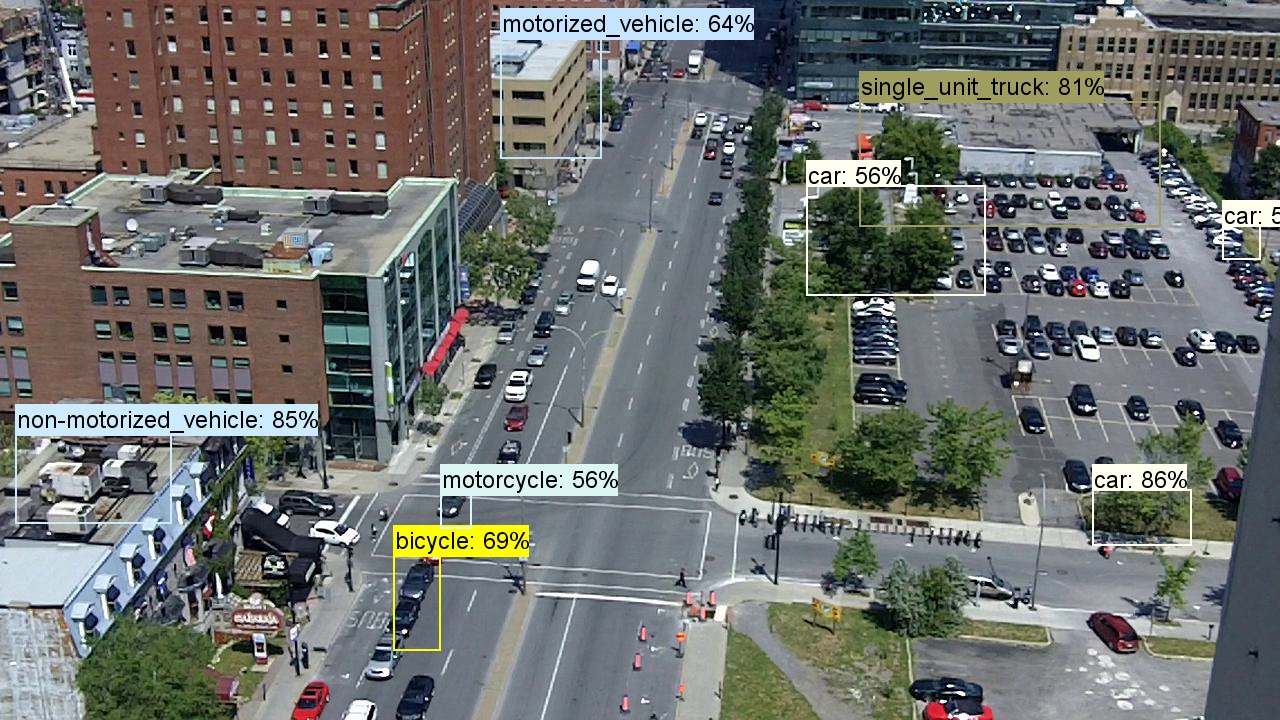}
\caption{Typical detections obtained from the Ren\'e-L\'evesque video.} \label{fig:rene}
\end{figure}

Table~\ref{tab_comparison} summarizes the results obtained with the baseline tracker. First of all, we were not able to obtain interesting results on the Ren\'e-L\'evesque video. From the evaluation, it is observed that the size of the objects greatly influences the performance of the proposed method because of the shortcomings of RFCN. When the size of the road users is too small, there are not enough details for the detector to distinguish the different types of objects reliably. Mis-detections are common in such cases, as observed in video Ren\'e-L\'evesque, for example in Figure~\ref{fig:rene}. Since the frames are captured at a higher altitude than the other urban scenes, the object detector has difficulties in detecting and classifying the objects clearly due to the lack of details. On the other hand, larger objects such as buildings have the tendency of being detected as they share similarities with the features learned by the detector. 

\begin{table}[]
\centering
\caption{Comparison of MOTA and MOTP scores for three videos of the Urban Tracker dataset with the inclusion and exclusion of label cost in the data association (the best results are in boldface).} 
\label{tab_comparison}
\begin{tabular}{p{2cm}  p{1.6cm}  p{2.1cm} p{2.1cm}  p{2.1cm} p{2.1cm}}
\hline
\multirow{2}{*}{} & \multirow{2}{*}{\shortstack{No. \\ objects}} & \multicolumn{2}{c}{MOTP} & \multicolumn{2}{c}{MOTA} \\
                  &                                    & with labels            & without labels          & with labels             & without labels            \\
 \hline
Rouen             & 16                                 & 0.6870     & \bf{0.6893} & \bf{-0.1877}          & -0.4176          \\
Sherbrooke        & 20                                 & \bf{0.7488}     & 0.7324 & \bf{0.0266}          & -0.0023         \\
St-Marc           & 28                                 & \bf{0.7234}     & 0.7136 & -0.3657          & \bf{-0.2749}    \\
\hline
\end{tabular}
\end{table}



\begin{figure}
\centering
\includegraphics[width=0.6\linewidth]{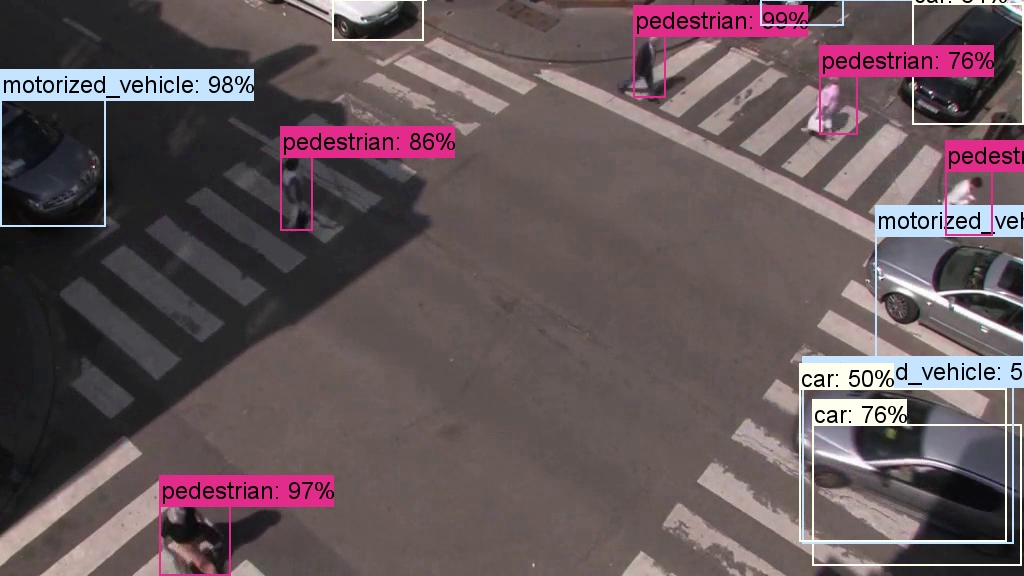}
\caption{An example of the redundant detection output for the same object.} \label{fig:redundant}
\end{figure}

Secondly, it can be noticed from Table~\ref{tab_comparison} that the MOTA results are negative and disappointing. This comes from the difficulty of interpreting the detections of RFCN. The same object is sometimes detected as several instances from the object detection module, as shown in Figure \ref{fig:redundant}. This often causes confusion and unnecessarily increases the number of detected objects and degrades the reported tracking performance. When there are no consecutive redundant detections, these redundant instances of the same object will usually be removed after a few frames since the object assignments are exclusive. 

Furthermore, contrarily to background subtraction or optical flow-based methods, RFCN gives detection outputs also for cars that are parked or for a car on a advertising billboard. Therefore, the data association process is distracted by many irrelevant objects. In such cases, standard NMS is not very useful in a traffic scene. 
Although NMS is used, it is insufficient to eliminate all the redundancies.

Since the proposed method is intrinsically dependent on the results from the detection module, the mis-detections propagate and deteriorate the overall MOT performance. In this case, the existence of redundant tracks severely affects the MOTA score such that it falls into the negative range, as shown in Table~\ref{tab_comparison}. The MOTA takes into account the number of misses, false positives and mismatches from the produced trajectories. 

However, it can be noted that MOT with inclusion of classification label generally gives higher MOTA. Among the different classes of labels from the detection module, the non-motorized vehicle label is currently excluded in the tracking framework since the occurrence of non-motorized vehicles is very rare in this experiment while parts of the background are sometimes mistakenly identified as objects from this class. MOTP is sometimes slightly better without labels as there are cases where tracking of an object fails in successive frames due to the switch of labels from the detection results. This is because with the labels, some matches are penalized and rejected because they are higher than the cost threshold. Therefore, the total number of matches is different, leading to slightly different values for MOTP. This occurrence is common among classes that share similarities such as pedestrians, bicycles and motorcycles, resulting in redundant tracks or fragmented tracks for the same object and thus lowering the overall tracking performance. 

\section{Conclusion}
In this paper, the use of a modern multiclass object detector was investigated for the MOT task in traffic scenes. It was integrated in a baseline multiple object tracker. Results show that classification labels can be beneficial in MOT. However, the outputs of the multiclass object detector are hardly usable because they include a large number of false detections, or detections of objects that are not of interest in the current application (e.g. parked cars). Small objects are also difficult to detect. As a result, to use such a detector, its output needs to be combined with another detector that can focus more precisely on objects of interest such as background subtraction or optical flow. 

\section{Acknowledgement}
This research is partly funded by Fonds de Recherche du Quebec -Nature et Technologies(FRQ-NT) with team grant No. 2016-PR-189250 and Polytechnique Montr\'{e}al PhD Fellowship. We gratefully acknowledge the support of NVIDIA Corporation with the donation of the Titan X GPU used for this work.
%
%
%

\begin{thebibliography}{99}
\bibitem{ref_miotcd}
Luo, Zhiming and Frederic, B and Lemaire, Carl and Konrad, Janusz and Li, Shaozi and Mishra, Akshaya and Achkar, Andrew and Eichel, Justin and Jodoin, Pierre-Marc and others: MIO-TCD: A new benchmark dataset for vehicle classification and localization. IEEE Transactions on Image Processing\ (2018)

\bibitem{ref_rfcn}
Dai, Jifeng and Li, Yi and He, Kaiming and Sun, Jian: R-fcn: Object detection via region-based fully convolutional networks. In: Advances in neural information processing systems, pp. 379--387. (2016)

\bibitem{ref_urbantracker}
Jodoin, Jean-Philippe and Bilodeau, Guillaume-Alexandre and Saunier, Nicolas: Tracking all road users at multimodal urban traffic intersections. IEEE Transactions on Intelligent Transportation Systems \textbf{17}(11), 99--110 (2016)

\bibitem{ref_sort}
Bewley, Alex and Ge, Zongyuan and Ott, Lionel and Ramos, Fabio and Upcroft, Ben: Simple online and realtime tracking. In: 2016 IEEE International Conference on Image Processing (ICIP)
on Proceedings, pp. 3464-3468. (2016)

\bibitem{ref_hungarian}
Kuhn, Harold W: The Hungarian method for the assignment problem. Naval research logistics quarterly \textbf{2}(5), 83--97 (1955)

\bibitem{ref_clearmot}
Bernardin, Keni and Stiefelhagen, Rainer: Evaluating multiple object tracking performance: the CLEAR MOT metrics. Journal on Image and Video Processing (2008)

\bibitem{ref_ensemble}
Malisiewicz, Tomasz and Gupta, Abhinav and Efros, Alexei A: Ensemble of exemplar-svms for object detection and beyond. In: 9Computer Vision (ICCV), 2011 IEEE International Conference, pp. 89--96. Publisher, Location (2011)

\bibitem{ref_objdetection}
Felzenszwalb, Pedro F and Girshick, Ross B and McAllester, David and Ramanan, Deva: Object detection with discriminatively trained part-based models. In: IEEE transactions on pattern analysis and machine intelligence, pp. 1627--1645. (2010)

\bibitem{ref_kalman}
Kalman, Rudolph Emil: A new approach to linear filtering and prediction problems. Journal of basic Engineering \textbf{82}(1), 35--45 (1960)

\bibitem{ref_mendes15vehicle}
Mendes, Jean Carlo and Bianchi, Andrea Gomes Campos and J{\'u}nior, Alvaro R Pereira: Vehicle Tracking and Origin-Destination Counting System for Urban Environment. In: Proceedings of the International Conference on Computer Vision Theory and Applications, (2015)  

\bibitem{ref_saunier_feature_based_2006}
Saunier, N. and Sayed, T.: A feature-based tracking algorithm for vehicles in intersections. In: Saunier, N. and Sayed, T.: A feature-based tracking algorithm for vehicles in intersections. In: The 3rd Canadian Conference on Computer and Robot Vision, pp. 59--59. (2006)

\bibitem{ref_kim_real_2008}
Kim, {ZuWhan}: Real time object tracking based on dynamic feature grouping with background subtraction, In: {IEEE} Conference on Computer Vision and Pattern Recognition (CVPR), pp. 1-8. (2008) 

\bibitem{ref_jun_tracking_2008}
Jun, Goo and Aggarwal, J. K. and Gokmen, Muhittin: Tracking and Segmentation of Highway Vehicles in Cluttered and Crowded Scenes. In: Proceedings of the 2008 {IEEE} Workshop on Applications of Computer Vision, pp. 1-6. (2008)

\bibitem{ref_coifman_real-time_1998}
Benjamin Coifman and David Beymer and Philip McLauchlan and Jitendra Malik: A real-time computer vision system for vehicle tracking and traffic surveillance. In: Transportation Research Part C: Emerging Technologies, pp. 271--288. (1998)

\bibitem{ref_KLT}
Shi, J. and Tomasi, C.: Good features to track. In: Computer Vision and Pattern Recognition. pp. 593--600. (1994)
 
\bibitem{ref_Torabi2009}
Torabi, A. and Bilodeau, G. -A: A Multiple Hypothesis Tracking Method with Fragmentation Handling. In: Canadian Conference on Computer and Robot Vision. pp. 8--15. (2009)

\bibitem{ref_Beymer1997}
Beymer, D. and McLauchlan, P. and Coifman, B. and Malik, J.: A real-time computer vision system for measuring traffic parameters. In: IEEE Computer Society Conference on Computer Vision and Pattern Recognition. pp. 495--501. (1997)

\bibitem{ref_Fuentes2006}
Luis M. Fuentes and Sergio A. Velastin: People tracking in surveillance applications. In: Image and Vision Computing. pp. 1165--1171. (2006)

\bibitem{ref_Aslani2013}
Sepehr Aslani and  Homayoun Mahdavi-Nasab: Optical Flow Based Moving Object Detection and Tracking for Traffic Surveillance. In: International Journal of Electrical, Robotics, Electronics and Communications Engineering. pp. 773--777. (2013)

\bibitem{ref_JodoinITS2016}
J. P. Jodoin and G. A. Bilodeau and N. Saunier: In: IEEE Winter conference on Applications of Computer Vision. pp. 885-892.  (2016) 

\bibitem{ref_yolo}
Redmon, Joseph and Divvala, Santosh and Girshick, Ross and Farhadi, Ali: You only look once: Unified, real-time object detection. In: Proceedings of the IEEE conference on computer vision and pattern recognition, pp. 779--788. (2016)





\end{thebibliography}
%

\end{document}